\documentclass{article}
\usepackage{smlia2024}
\usepackage{multicol}
\usepackage{amsmath}
\usepackage[utf8]{inputenc} 
\usepackage[T1]{fontenc}    
\usepackage{hyperref}       
\hypersetup{colorlinks,allcolors=blue}
\usepackage{url}            
\usepackage{booktabs}       
\usepackage{amsfonts}       
\usepackage{nicefrac}       
\usepackage{microtype}      
\usepackage{cleveref}       
\usepackage{lipsum}         
\usepackage{graphicx}
\usepackage[numbers]{natbib}
\usepackage{doi}
\usepackage{wrapfig}
\usepackage{float}
\usepackage{xcolor}

\title{Towards Foundation Models for the Industrial Forecasting of Chemical Kinetics}
\renewcommand{\shorttitle}{Towards Forecasting FMs for Industrial Utility}

\date{}

\newif\ifuniqueAffiliation
\uniqueAffiliationtrue

\ifuniqueAffiliation 

\usepackage{authblk}

\author[1,3]{%
	Imran Nasim\thanks{\texttt{imran.nasim@ibm.com}, \texttt{i.nasim@surrey.ac.uk}}}%
\author[2]{%
	Jo\~ao Lucas de Sousa Almeida\thanks{\texttt{joao.lucas.sousa.almeida@ibm.com}}}%
\affil[1]{IBM, UK}
\affil[2]{IBM Research Brazil}
\affil[3]{Department of Mathematics, University of Surrey, Guildford, GU2 7XH, Surrey, UK}
\fi
\hypersetup{
pdftitle={A template for the arxiv style},
pdfsubject={q-bio.NC, q-bio.QM},
pdfauthor={David S.~Hippocampus, Elias D.~Striatum},
pdfkeywords={First keyword, Second keyword, More},
}

\begin{document}
\maketitle
\begin{abstract}
Scientific Machine Learning is transforming traditional engineering industries by enhancing the efficiency of existing technologies and accelerating innovation, particularly in modeling chemical reactions. Despite recent advancements, the issue of solving stiff
chemically reacting problems within computational fluid dynamics remains a significant issue. In this study we propose a novel approach utilizing a multi-layer-perceptron mixer architecture (MLP-Mixer) to model the time-series of stiff chemical kinetics. We evaluate this method using the ROBER system, a benchmark model in chemical kinetics, to compare its performance with traditional numerical techniques. This study provides insight into the industrial utility of the recently developed MLP-Mixer architecture to model chemical kinetics and provides motivation for such neural architecture to be used as a base for time-series foundation models. 
\end{abstract}

\begin{multicols}{2}
\twocolumn
\section{Introduction}
Scientific Machine Learning (SciML) has become a valuable approach for tackling the numerically challenging and computationally demanding problems often found in real-world industrial cases, providing reliable solutions by leveraging large-scale industrial data.
An exciting area of significant industrial impact is in modelling chemical reactions \citep{goswami2024_chemical}.
The study of chemical reactions is fundamentally important across various industries, providing insights crucial for innovation, product quality, and environmental management. This has led to significant efforts in developing specific optimization strategies \citep{murray2016_chemical_application,shields2021_bayesian,taylor2023_chemical}. However, modelling chemical reactions are non trivial and often requires you to solve stiff ordinary differential equations (ODEs) within Computational Fluid Dynamics (CFD) simulations \citep{aditya2019_direct}.
Stiff chemical kinetics are computationally expensive to solve within CFD simulations which has prompted various deep learning based techniques to solve such problems including ResNet \cite{brown2021_novel_resnet}, Physics informed Neural Network \citep{de2022_pinn}, Autoencoders \citep{aditya2019_direct} and Deep Operator Networks \citep{goswami2024_chemical} to model chemical reactions. Despite some promising results from the aforementioned SciML methods, the issue of solving such stiff
chemically reacting problems within the CFD framework remains a significant issue. In this study we propose a novel approach to learning stiff chemical kinetics using an MLP-Mixer method named PatchTSMixer \cite{Ekambaram_2023}, a promising base architecture for time-series foundation models. We test this method on the Robertson (ROBER) system \citep{robertson1966solution}, which is highly prominent in the field of chemical kinetics and often used as a standard for evaluating and comparing different numerical techniques. 
Our results provide evidence towards the industrial utility of the MLP-Mixer architecture for modeling and forecasting chemical kinetics.

\section{Models and Experiments}\label{sec:mod_exp}
\begin{figure}[H]
    \centering    
    \includegraphics[scale=0.15]{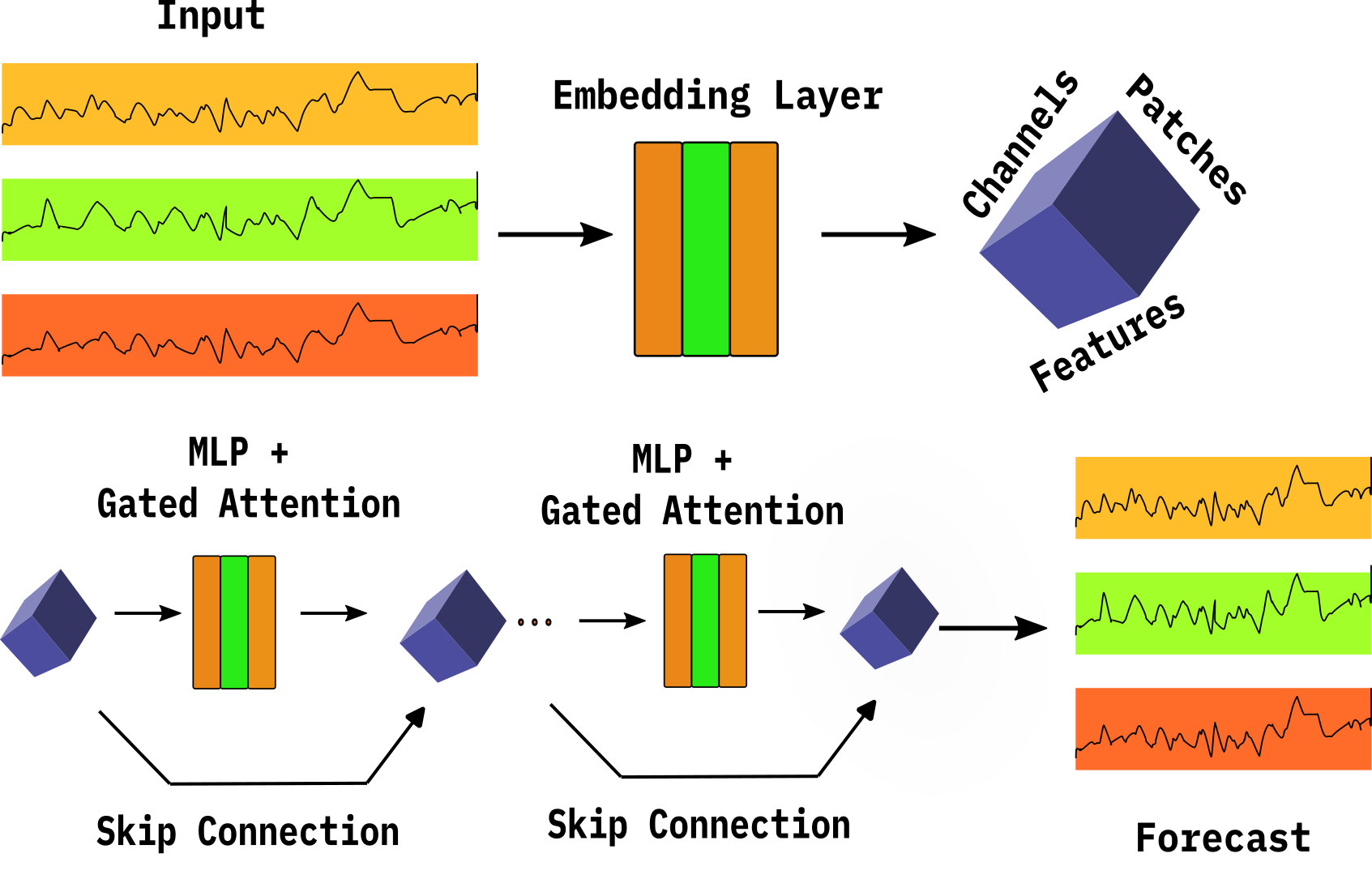}
    \caption{Schematic of the MLP-Mixer method PatchTSMixer.}
    \label{fig:patchtsmixer_schematic}
\end{figure}

\textbf{Chemical Kinetic System.}
The ROBER system is a nonlinear coupled ODE system designed to model the chemical kinetics of an autocatalytic reaction, see Eq \eqref{eq:system}. 
As there is a large discrepancy between the constants $k_1$, $k_2$ and $k_3$, the system has a tendency to present considerable stiffness, making the
numerical integration nontrivial.\newline
\textbf{MLP-Mixer Method.} 
PatchTSMixer is a relatively novel MLP-Mixer architecture intended to provide a lightweight alternative to perform time-series forecasting. A schematic of the architecture is given in Fig \ref{fig:patchtsmixer_schematic}. 
The first PatchTSMixer stage splits multivariate time-series into small patches which are then transformed by an intermediary layer into a multi-dimensional tensor. 
This multi-dimensional tensor is subsequently passed through multiple MLP mixer layers, each one learning correlations between patches and channels. At the last stage, the embedding is reshaped in order to produce the final output.
The model can be trained in two ways during the pretraining stage: as a masked autoencoder, which is suitable for multiple downstream finetuning tasks, and for direct forecasting, the approach preferred in this work.
PatchTSMixer is able of predicting a number $h$ of future timesteps given a context length of size $H$, which can be optimized in hyperparameter tuning.\newline
\textbf{Data and Training.}
The dataset used to perform this experiments was generated using the LSODA solver, which is designed to solve stiff problems. The ROBER system was integrated for $t=10^5\,s$, from a randomly chosen initial state, $[0.776, 6.913\,10^{-5},0,081]$ and recorded with resolution of $\Delta\,t=1\,s$. This initial condition was chosen to produce the stiff chemical kinetics.
The dataset is split as $60\%$ for training, $20\%$ for validation and $20\%$ for testing. We have trained PatchTSMixer using a context length $H=512$, a prediction horizon $h=100$ and a patch length $p=8$. The model was trained for $300$ epochs using the AdamW optimizer. After the training stage, the model was used to forecast the testing region by small intervals in order to create Fig \ref{fig:forecast_rober_plot}. All the training and tests were performed in a single 20-core Intel Xeon CPU.

\section{Results}

\begin{figure}[H]
    \centering    
    \includegraphics[scale=0.49]{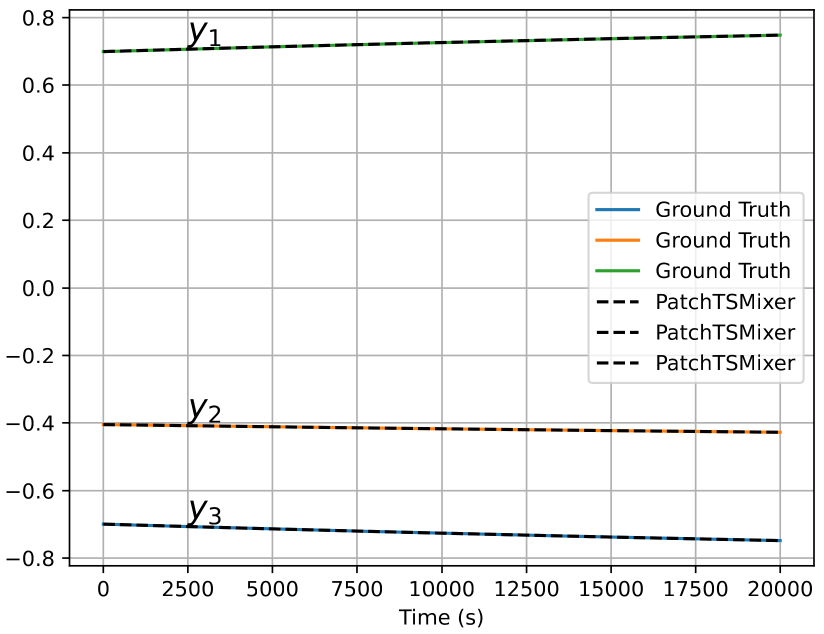}
    \caption{Batchwise extrapolation using PatchTSMixer. The variables were normalized in order to enhance the visualization.}
    \label{fig:forecast_rober_plot}
\end{figure}

To determine if PatchTSMixer can accurately capture and forecast stiff chemical kinetics, we compare its extrapolation predictions with those of the ground truth evolution obtained from the stiff numerical solver. This comparison is presented in Fig \ref{fig:forecast_rober_plot} for the final 20,000 timesteps of the evolution. We find that the extrapolation predictions of the chemical species $y_i$ obtained from PatchTSMixer both quantitatively and qualitatively match the evolution from the numerical solver. We evaluated the relative error for each extrapolation batch, corresponding to forecasts of 100 time steps, and found a mean error of $0.0166\%$ with a standard deviation of $0.0008\%$. We observed that PatchTSMixer is not so robust when used in dynamic extrapolation (when we use the output of a forecast batch to feed the next one) and it starts to deviate. However, it is possible that it can be enhanced with better choices of context and forecast dimensions. In Fig \ref{fig:forecast_rober_plot}, we plot the curve composed of the batchwise forecasts. It is possible to verify that even though there is an apparent smoothing, PatchTSMixer actually produces outputs with very small fluctuations, not visible at the scale of the plots. These characteristics persist for the additional tests, see Apx. \ref{appendix_b}.
We anticipate that this issue can be alleviated through a more in-depth investigation of its hyperparameters. These results provide evidence that PatchTSMixer can accurately capture and forecast stiff chemical kinetics and motivate the use of the MLP-Mixer method as a base architecture for time-series forecasting foundation models.

\section{Conclusion}
In this study we applied a recently proposed MLP-Mixer architecture, PatchTSMixer, in a novel SciML context to model the time-series evolution of stiff chemical kinetics which is of significant importance in industry. We evaluate our approach using the ROBER system, a benchmark model in chemical kinetics which describes the kinetics of an autocatalytic reaction. We find that there is a very strong agreement between the evolution of the chemical species forecasted by PatchTSMixer method compared to the ground truth. This result highlights the industrial utility of using an MLP-Mixer for forecasting the time-series evolution of stiff chemical kinetics and provides motivation for such architecture to be considered as a building block for modern day time-series foundation models.


\bibliographystyle{plain}
\bibliography{references} 

\begin{thebibliography}{1}

\bibitem{aditya2019_direct}
Konduri Aditya, Andrea Gruber, Chao Xu, Tianfeng Lu, Alex Krisman, Mirko~R Bothien, and Jacqueline~H Chen.
\newblock Direct numerical simulation of flame stabilization assisted by autoignition in a reheat gas turbine combustor.
\newblock {\em Proceedings of the Combustion Institute}, 37(2):2635--2642, 2019.

\bibitem{brown2021_novel_resnet}
Thomas~S Brown, Harbir Antil, Rainald L{\"o}hner, Fumiya Togashi, and Deepanshu Verma.
\newblock Novel dnns for stiff odes with applications to chemically reacting flows.
\newblock In {\em High Performance Computing: ISC High Performance Digital 2021 International Workshops, Frankfurt am Main, Germany, June 24--July 2, 2021, Revised Selected Papers 36}, pages 23--39. Springer, 2021.

\bibitem{de2022_pinn}
Mario De~Florio, Enrico Schiassi, and Roberto Furfaro.
\newblock Physics-informed neural networks and functional interpolation for stiff chemical kinetics.
\newblock {\em Chaos: An Interdisciplinary Journal of Nonlinear Science}, 32(6), 2022.

\bibitem{Ekambaram_2023}
Vijay Ekambaram, Arindam Jati, Nam Nguyen, Phanwadee Sinthong, and Jayant Kalagnanam.
\newblock Tsmixer: Lightweight mlp-mixer model for multivariate time series forecasting.
\newblock In {\em Proceedings of the 29th ACM SIGKDD Conference on Knowledge Discovery and Data Mining}, KDD ’23. ACM, August 2023.

\bibitem{goswami2024_chemical}
Somdatta Goswami, Ameya~D Jagtap, Hessam Babaee, Bryan~T Susi, and George~Em Karniadakis.
\newblock Learning stiff chemical kinetics using extended deep neural operators.
\newblock {\em Computer Methods in Applied Mechanics and Engineering}, 419:116674, 2024.

\bibitem{murray2016_chemical_application}
Paul~M Murray, Fiona Bellany, Laure Benhamou, Dejan-Kre{\v{s}}imir Bu{\v{c}}ar, Alethea~B Tabor, and Tom~D Sheppard.
\newblock The application of design of experiments (doe) reaction optimisation and solvent selection in the development of new synthetic chemistry.
\newblock {\em Organic \& biomolecular chemistry}, 14(8):2373--2384, 2016.

\bibitem{robertson1966solution}
HH~Robertson.
\newblock The solution of a set of reaction rate equations.
\newblock {\em Numerical analysis: an introduction}, 178182, 1966.

\bibitem{shields2021_bayesian}
Benjamin~J Shields, Jason Stevens, Jun Li, Marvin Parasram, Farhan Damani, Jesus I~Martinez Alvarado, Jacob~M Janey, Ryan~P Adams, and Abigail~G Doyle.
\newblock Bayesian reaction optimization as a tool for chemical synthesis.
\newblock {\em Nature}, 590(7844):89--96, 2021.

\bibitem{taylor2023_chemical}
Connor~J Taylor, Alexander Pomberger, Kobi~C Felton, Rachel Grainger, Magda Barecka, Thomas~W Chamberlain, Richard~A Bourne, Christopher~N Johnson, and Alexei~A Lapkin.
\newblock A brief introduction to chemical reaction optimization.
\newblock {\em Chemical Reviews}, 123(6):3089--3126, 2023.

\end{thebibliography}

\appendix
\section{Chemical Kinetic System}\label{appendix}
The ROBER system can be written as a system of non-linear coupled ODE, which is presented in Eq.\eqref{eq:system}.  The system models a set of reactions where species $y_1$, $y_2$ and $y_3$ interact. The rates of these reactions are governed by the rate constants $k_1$, $k_2$ and $k_3$.
\begin{align}
&\frac{dy_1}{dt} = -k_1\,y_1 + k_3 y_2 y_3 \nonumber\\
&\frac{dy_2}{dt} = k_1\,y_1 - k_2\,y_2^2 - k_3\,y_2\,y_3 \label{eq:system}\\
&\frac{dy_3}{dt} = k_2\,y_2^2 \nonumber
\end{align}

\section{Additional Numerical Tests}\label{appendix_b}
\begin{figure}[H]
    \centering    
    \includegraphics[scale=0.5]{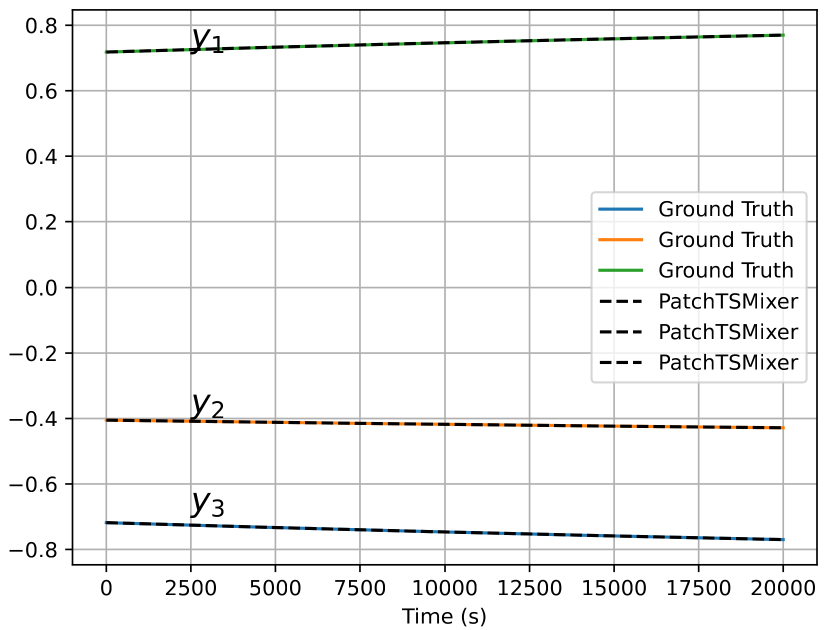}
    \centering    
    \includegraphics[scale=0.5]{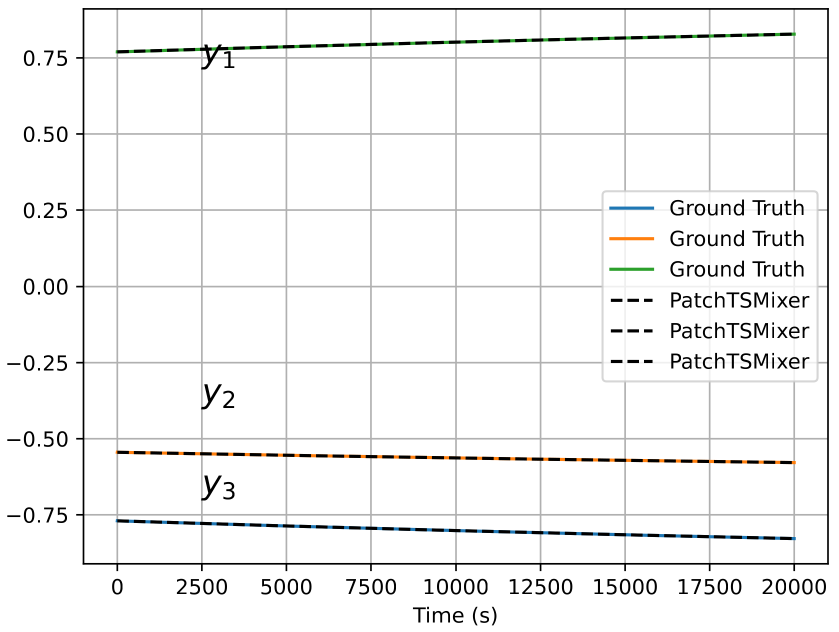}
    \caption{The batchwise extrapolation performed for two slightly different test cases.}
    \label{fig:forecast_rober_plot_extended}
\end{figure}

The data for the additional test cases seen in Fig \ref{fig:forecast_rober_plot_extended} were generated using the initial conditions $[0.879, 7.816\,10^{-5}\,0.077]$ and $[0.693, 4.254\,10^{-5}\, 0.390]$. This initial conditions we chosen as they produce the stiff chemical kinetics behaviour.

\end{multicols}
\end{document}